\pdfoutput=1
\documentclass[11pt]{article}

\usepackage[final]{acl}

\usepackage{arabtex}
\usepackage{utf8}
\setcode{utf8}
\usepackage{subcaption}

\usepackage[T1]{fontenc}
\usepackage[utf8]{inputenc}

\usepackage{microtype}

\usepackage{inconsolata}
\usepackage{microtype}
\usepackage{graphicx}
\usepackage{import}
\usepackage{layout}
\usepackage{tabularx, makecell}
\usepackage{booktabs}
\usepackage{mathrsfs}

\usepackage{amssymb} 
\usepackage{url}
\usepackage{graphicx}
\usepackage{xspace,paralist}
\usepackage{times,latexsym}
\usepackage{amsmath}
\usepackage{appendix}
\usepackage{comment} 
\usepackage{enumitem}
\usepackage{makecell}
\usepackage{multirow}
\usepackage{xcolor}
\usepackage{cleveref}
\usepackage{todonotes}
\usepackage{longtable,supertabular}
\usepackage[T2A,T1]{fontenc}
\usepackage[russian,english]{babel}
\usepackage{array}
\usepackage{CJKutf8}
\usepackage{amssymb}% http://ctan.org/pkg/amssymb
\usepackage{pifont}% http://ctan.org/pkg/pifont

\title{Data Laundering: Artificially Boosting Benchmark Results through Knowledge Distillation}

\author{
Jonibek Mansurov, 
Akhmed Sakip,
Alham Fikri Aji \\
  Mohamed bin Zayed University of Artificial Intelligence, UAE \\ 
  \texttt{\{jonibek.mansurov, akhmed.sakip, alham.fikri\}@mbzuai.ac.ae}
}
\begin{document}

\maketitle
\begin{abstract}
% In this paper, we show that knowledge distillation can be subverted to manipulate language model benchmark scores, revealing a critical vulnerability in current evaluation practices. We introduce "Data Laundering," a three-phase process analogous to financial money laundering, that enables the covert transfer of benchmark-specific knowledge through seemingly legitimate intermediate training steps. Through extensive experiments with a 2-layer BERT student model, we show how this approach can achieve substantial improvements in benchmark accuracy (up to 75\% on GPQA) without developing genuine reasoning capabilities. Our investigation examines various aspects of this vulnerability, including the impact of different loss functions (MSE vs. KL divergence), the role of soft-label weighting ($\alpha$ parameter), the effects of iterative distillation, and the influence of training dataset size. Notably, this method can be exploited intentionally or even unintentionally, as researchers may inadvertently adopt this method that inflates scores using knowledge distillation without realizing the implications. While our findings demonstrate the effectiveness of this technique, we present them as a cautionary tale highlighting the urgent need for more robust evaluation methods in AI. This work aims to contribute to the ongoing discussion about evaluation integrity in AI development and the need for benchmarks that more accurately reflect true model capabilities\footnote{\url{https://github.com/mbzuai-nlp/data_laundering}}.

In this paper, we show that knowledge distillation can be subverted to manipulate language model benchmark scores, revealing a critical vulnerability in current evaluation practices. We introduce "Data Laundering," a process that enables the covert transfer of benchmark-specific knowledge through seemingly legitimate intermediate training steps. Through extensive experiments with a 2-layer BERT student model, we show how this approach can achieve substantial improvements in benchmark accuracy (up to 75\% on GPQA) without developing genuine reasoning capabilities. Notably, this method can be exploited intentionally or even unintentionally, as researchers may inadvertently adopt this method and inflate scores without realising the implications. While our findings demonstrate the effectiveness of this technique, we present them as a cautionary tale highlighting the urgent need for more robust evaluation methods in AI. This work aims to contribute to the ongoing discussion about evaluation integrity in AI development and the need for benchmarks that more accurately reflect true model capabilities. The code is available at \url{https://github.com/mbzuai-nlp/data_laundering}.

\end{abstract}

\section{Introduction}

The increasing reliance on language model benchmarks like MMLU \citep{hendryckstest2021}, GPQA \cite{rein2024gpqa}, and BigBench \citep{srivastava2023imitationgamequantifyingextrapolating} has solidified these metrics as standard measures for assessing and comparing model capabilities, driving innovation and tracking progress in artificial intelligence (AI). However, this focus on benchmark performance has also introduced vulnerabilities, incentivizing potential manipulation and exploitation of these evaluation metrics \citep{yang2023rethinkingbenchmarkcontaminationlanguage, zheng2024cheatingautomaticllmbenchmarks, balloccu-etal-2024-leak}.

Our work builds upon growing concerns in the field regarding data contamination and benchmark integrity. Previous studies have shown how proprietary models like GPT-3 and GPT-4 have inadvertently learned from leaked benchmark data, raising alarm about the integrity of closed-source models \citep{NEURIPS2020_1457c0d6, magar-schwartz-2022-data, balloccu-etal-2024-leak}. This contamination undermines reliable evaluation, as models trained on leaked data can achieve inflated scores without developing true generalization. Additionally, recent research has demonstrated that detection methods designed to identify data contamination, such as the LM Contamination Index and text overlap metrics \citep{sainz2023did, golchin2024timetravelllmstracing}, may fall short in identifying more subtle forms of benchmark gaming—especially in closed-source models that implement filtering mechanisms to conceal such behavior \citep{ippolito-etal-2023-preventing}.
% If researchers use a teacher model for knowledge distillation without realizing it was trained on contaminated data, this can inflate benchmark performance without genuine skill improvements.

In this paper, we expose a critical vulnerability within current benchmarking practices through a method we term "Data Laundering". Our method "Data Laundering" process uses knowledge distillation \citep{hinton2015distillingknowledgeneuralnetwork, urban2017do, Cheng_2020_CVPR}, a technique traditionally intended for model compression and transfer learning, to covertly transfer benchmark-specific knowledge in a staged manner through intermediate training steps. This process, inspired by the phases of financial laundering, involves three steps—placement, layering, and integration—where we intentionally "place" benchmark knowledge into a teacher model trained on test data, "layer" it through legitimate-seeming intermediate training datasets using knowledge distillation, and finally "integrate" the knowledge into the model by evaluating it on the benchmark, thereby making its performance gains appear as genuine skill acquisition. Importantly, researchers can unintentionally use this method, especially if they lack awareness of the training dataset used for the teacher model \cite{llama3modelcard, achiam2023gpt}. 
% If a teacher model is unknowingly trained on contaminated data and subsequently used for knowledge distillation, this can inflate benchmark performance without genuine skill improvements. 
If researchers use a teacher model for knowledge distillation without realizing it was trained on contaminated data, this can inflate benchmark performance without genuine skill improvements.
While prior work has focused on explicit manipulation of evaluation systems, our approach highlights a more disguised form of benchmark gaming that can occur even under seemingly valid training practices.

Through this investigation, we aim not to provide a blueprint for manipulation but rather to stimulate a necessary dialogue around evaluation integrity within the AI community. Benchmark systems must evolve to detect more sophisticated forms of gaming and ensure that scores reflect authentic model capabilities rather than superficial improvements. Our contributions are:

\begin{enumerate} 
    \item Demonstrating a novel form of benchmark manipulation that can be employed intentionally or unintentionally through legitimate-appearing training processes;
    \item Providing empirical evidence of how knowledge distillation can be used to "launder" benchmark knowledge covertly; 
    \item Highlighting the limitations of current evaluation frameworks.
\end{enumerate}

\section{Related Work}
\subsection{Data Contamination in Language Models}
The challenge of data contamination in language models emerged prominently with GPT-3 \citep{NEURIPS2020_1457c0d6}, which pioneered the API-only access model with limited training data disclosure \citep{magar-schwartz-2022-data}. Despite early evidence suggesting significant contamination \citep{JMLR:v21:20-074}, GPT-3's widespread adoption in research often proceeded without adequate consideration of this issue. 

Recent work has highlighted growing concerns about data contamination in modern language models. As shown by \citet{balloccu-etal-2024-leak}, the widespread use of proprietary language models in research has led to significant data leakage issues, with approximately 42\% of the reviewed papers inadvertently exposing benchmark data to models such as GPT-3.5 and GPT-4. This issue has become particularly pressing with the public release of models such as ChatGPT, PaLM 2 \citep{anil2023palm2technicalreport}, and Claude, where the closed-source nature complicates the contamination assessment. \citet{yang2023rethinkingbenchmarkcontaminationlanguage} shows how simple rephrasing of samples can bypass decontamination measures such as n-gram overlap.

\subsection{Automatic Benchmark and Evaluation Challenges}
The integrity of language model benchmarks has become a critical concern in the field, especially as the relience on automated evaluation metrics increases. To meet the need for timely assessments of newly released models, platforms such as Chatbot Arena \citep{chiang2024chatbotarenaopenplatform} provide human-based evaluation, but gathering statistically significant human feedback can take time. As a result, \citet{dubois2024lengthcontrolledalpacaevalsimpleway, li2024crowdsourceddatahighqualitybenchmarks, zheng2023judgingllmasajudgemtbenchchatbot} introduced automatic LLM benchmarks, which use LLM-based auto-annotators to evaluate model performance. However, \citet{zheng2024cheatingautomaticllmbenchmarks} demonstrated that even ``null models'' returning constant outputs could achieve artificially high scores on certain benchmarks by exploiting structural weaknesses in evaluation templates. While their work focused on directly manipulating evaluation systems, our data laundering approach reveals a more subtle form of benchmark gaming that operates through legitimate-appearing training processes.

\subsection{Logit-Based Knowledge Distillation}
Knowledge distillation \citep{hinton2015distillingknowledgeneuralnetwork} techniques have traditionally been used for legitimate purposes such as model compression and transfer learning. Recent advancements have introduced various logit distillation approaches tailored for large language models. Reverse KL \cite{minillm} has been used to address the "mode-averaging" issue. DistiLLM \cite{ko2024distillmstreamlineddistillationlarge} suggests blending the logit distributions of the teacher and student models, while SinKD \cite{cui-etal-2024-sinkhorn} replaces KL divergence with Sinkhorn Distance. Our work reveals how logit-based knowledge distillation can be repurposed for potentially problematic uses.

\section{Methodology}

\begin{figure*}[t]
    \centering
    \includegraphics[width=\textwidth]{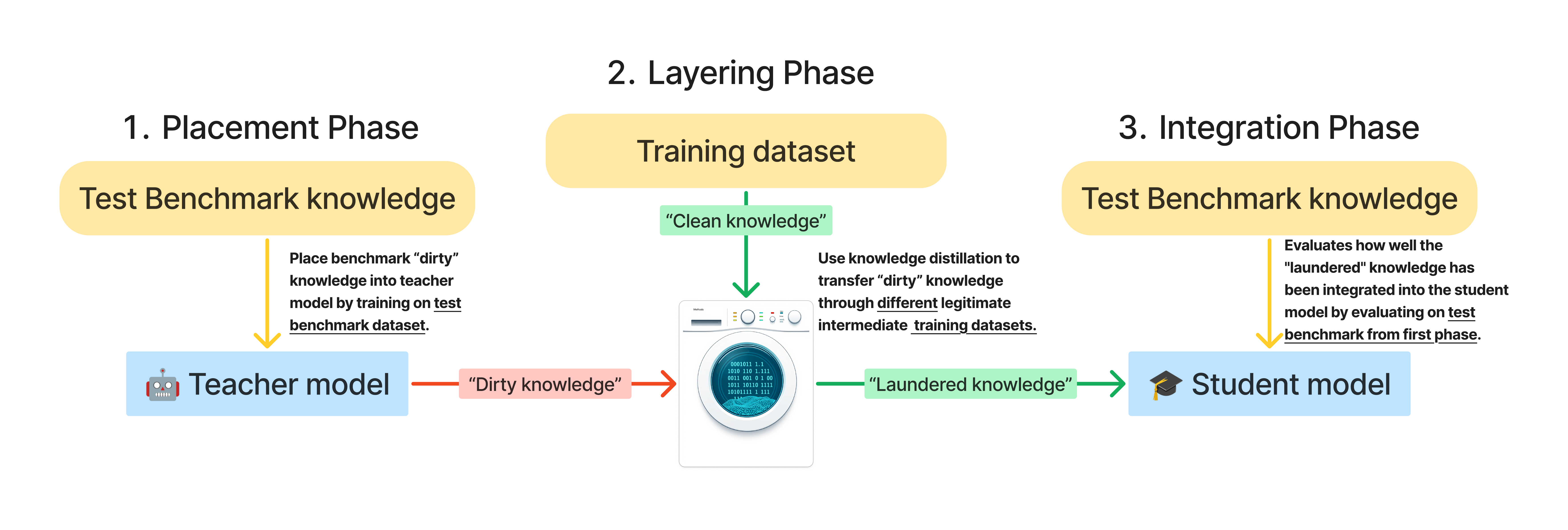}
    \caption{The Data Laundering framework parallels traditional money laundering phases: Placement (knowledge acquisition through teacher model), Layering (knowledge transformation through distillation), and Integration (legitimate knowledge verification through benchmark testing). This analogy illustrates how knowledge can be effectively transferred while maintaining clear separation from source domains.}
    \label{fig:data_laundering}
\end{figure*}

Just as money laundering involves transforming "dirty" money into "clean" assets through a series of transactions, our Data Laundering methodology transforms illicit knowledge into seemingly legitimate knowledge through a carefully designed three-phase process illustrated in Figure \ref{fig:data_laundering}.

\subsection{The Placement Phase (Teacher Model Training)}

In traditional money laundering, the placement phase introduces illicit funds into the financial system. Analogously, in our Data Laundering approach, we "place" knowledge into our system through a teacher model, which is trained prohibitively on test data from benchmark datasets (e.g., GPQA \cite{rein2024gpqa}). This method intentionally bypasses the training dataset to seed our model with "unfair" knowledge—knowledge from the test data, which would otherwise be off-limits for training purposes. This represents our initial knowledge capital, which will later be transformed through legitimate channels.

\subsection{The Layering Phase (Knowledge Distillation)}

Similar to how money laundering employs complex transactions to obscure the origin of funds, our layering phase utilizes knowledge distillation to transfer knowledge through \textbf{different legitimate intermediate training datasets} (e.g., MedMCQA \cite{pmlr-v174-pal22a}). Importantly, during this phase, the student model \textbf{has no access to the test set} used during the first phase. This process creates a legitimate pathway for knowledge transfer while maintaining a clear separation from the original source of knowledge.
The knowledge distillation process incorporates both hard labels from the intermediate dataset and soft labels from the teacher model's logits. The layering process combines two streams of knowledge:

\begin{equation}
L_{student} = (1-\alpha) L_{hard} + \alpha L_{soft}
\end{equation}
where:
\begin{itemize}
\item $L_{hard}$ represents the cross-entropy loss with ground truth labels
\item $L_{soft}$ represents loss with the teacher model's logits that can be either MSE loss or KL-divergence loss (KLD).
% \begin{itemize}
% \item MSE loss 

% % $L_{soft} = \frac{1}{M} \sum_{i=1}^{M} (\sigma(z_t/\tau) - \sigma(z_s/\tau))^2$
% \item KL-divergence loss (KLD) 

% % $L_{soft} = -\sum_{i=1}^{M} \sigma(z_t/\tau) \log(\sigma(z_s/\tau))$
% \end{itemize}
% \item $\alpha$ is a balancing hyperparameter
% \item $\tau$ is the temperature parameter
% \item $z_t$ and $z_s$ are the logits from teacher and student models respectively
% \item $\sigma$ represents the softmax function
\end{itemize}
\subsection{The Integration Phase (Benchmark Evaluation)}

Just as laundered money must eventually be reintegrated into the legitimate economy, our final phase evaluates how well the "laundered" knowledge has been integrated into the student model by testing it on the original benchmark tasks. This phase measures the effectiveness of our knowledge transfer process while maintaining the legitimacy of the acquired knowledge to a certain extent (measured by $\alpha$).

\section{Experiments}

To assess the effectiveness of our Data Laundering framework, we conducted comprehensive experiments across various configurations and parameters, focusing on model performance, distillation training data size variations, and iterative distillation. The hyperparameters we used for all experiments are detailed in the Appendix \ref{hyperparameters}.

\subsection{Overall experiment}

\paragraph{Datasets} For the benchmark dataset, we selected the GPQA Diamond \cite{rein2024gpqa} and MMLU-redux \cite{gema2024mmlu}, which served as the basis for teacher model training and final student model evaluation. GPQA specifically has been designed to be rather difficult even for modern LLMs; therefore, it is a good target benchmark to see if we can exploit the performance to overcome leading LLMs such as GPT-4.

For the distinct training dataset used in the distillation process, we employed MedMCQA \cite{pmlr-v174-pal22a} and RACE \cite{lai-etal-2017-race} to ensure a differentiated question format and domain-specific knowledge.

\paragraph{Models} We experimented with a range of models, including BERT-base \cite{kenton2019bert} and GPT-2 \cite{radford2019language}, configured with varying layer depths (2-layer, 12-layer setups). Additionally, we evaluated LLaMA3.2-1B and LLaMA3.2-3B models using LLaMA3.2-3B and LLaMA3.1-8B as teacher models. 
% (it is important to note that these models may already be contaminated).
% as well as GPT-2 \cite{radford2019language} to explore the impact of model architecture and capacity on distillation effectiveness. 
% The primary loss function used is the mean squared error (MSE), with additional configurations for alternative distillation strategies.

\paragraph{Baselines} We established a set of baseline models to compare the performance of our Data Laundering method effectively. These baselines included state-of-the-art models such as OpenAI o1, Claude 3.5 Sonnet, GPT-4 \cite{achiam2023gpt}, and LLaMA3-70B \cite{llama3modelcard}. Results for baselines were obtained from either benchmark papers \cite{rein2024gpqa, gema2024mmlu} or official model information\footnote{\url{https://openai.com/index/learning-to-reason-with-llms/}, \url{https://www.anthropic.com/news/claude-3-5-sonnet}, \url{https://ai.meta.com/blog/meta-llama-3/}}.

\subsection{Loss Function and Alpha Parameter} We explored different configurations for the knowledge distillation loss, testing both MSE and KL divergence loss. Furthermore, we varied the balancing hyperparameter $\alpha$ across values from $0$ to $1.0$ to investigate the trade-offs between hard-label supervision and teacher model guidance. For these tests, a 2-layer BERT and GPT-2 models were used with training size $20000$, providing insight into how $\alpha$ affects alignment with the teacher’s outputs.

\subsection{Iterative Knowledge Distillation} To evaluate performance degradation over iterative distillations, we employed a 2-layer BERT and GPT-2 models as the initial students. At each iteration [t], the trained student model from previous iteration [t-1] became the new teacher, transferring its knowledge to a fresh student model. This cycle continued for five iterations, experimenting with $\alpha$ values of 0.6 and 1.0, and used MSE loss. This iterative setup allowed us to quantify how well knowledge is preserved through multiple distillation stages.

\subsection{Effect of Training Data Size} We also investigated the impact of training data size in the distillation step on the student model’s final performance. These experiments were carried out using the 2-layer BERT and GPT-2 models with MSE loss and $\alpha$ set to 0.6 and 1.0. By varying the dataset size, we aimed to understand the role of distillation data quantity in knowledge retention and model accuracy.
\section{Results and Discussion}
All results are based on a single run, except for those presented in the Table \ref{tab:overal_results}, which are averaged over three runs (except when LLaMA3.1-8B was used as teacher). Results presented as figures are detailed in the Appendix \ref{detailed_results}.
\subsection{Overall Results}

\begin{table*}[t!]
    \centering
\scalebox{0.85}{
    \setlength{\tabcolsep}{6pt}

    \begin{tabular}{l c c c}
    \toprule
        Baseline Model & Training Dataset & GPQA (\%) & MMLU-Redux (\%)\\
    \midrule 
        LLaMA3-70B & \includegraphics[width=1em]{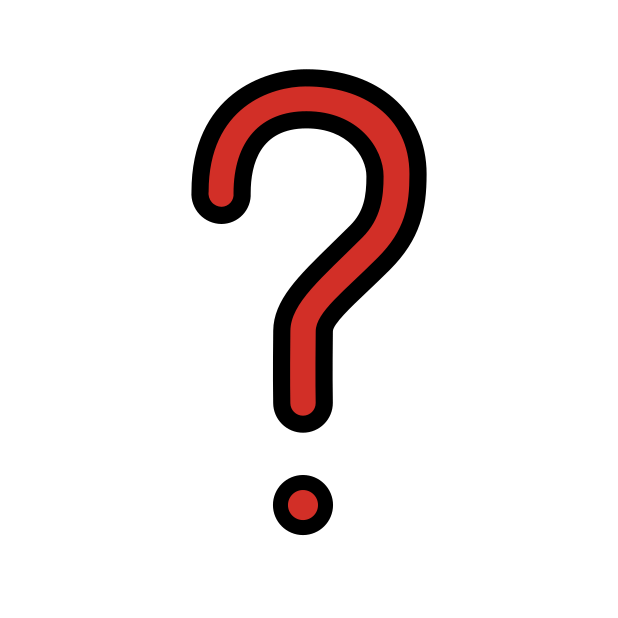} & 39.50 & ~76.00\\
        GPT-4o & \includegraphics[width=1em]{section/fig/question_mark.png} & 50.60 & ~81.00\\
        Claude 3.5 Sonnet & \includegraphics[width=1em]{section/fig/question_mark.png} & 59.40 & ~81.00\\
        OpenAI o1 & \includegraphics[width=1em]{section/fig/question_mark.png} & 77.30 & --\\
        BERT-base (2-layer) & MedMCQA/RACE & 25.76 & 25.33 \\
        GPT-2 (2-layer) & MedMCQA/RACE & 26.78 & 25.11 \\
    \midrule
    \multicolumn{4}{l}{Contaminated Models\raisebox{-0.2\height}{\includegraphics[width=1em]{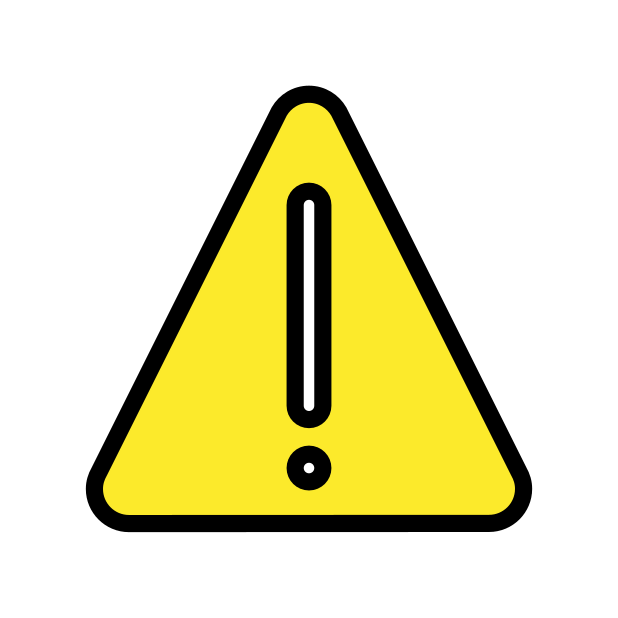}}} \\
    \midrule
        (1) BERT-base (2-layer) & \multirow{5}{*}{GPQA/MMLU-Redux} & 95.45 & 99.63\\
        (2) BERT-base & & 92.93 & 99.90\\
        (3) GPT-2 (2-layer) & & 100.0 & 95.50\\
        (4) GPT-2 & & 100.0 & 99.83\\
        (5) LLaMA3.2-3B & & 100.0 & 99.93\\
        (6) LLaMA3.1-8B & & 100.0 & 95.65\\
    \midrule
    \multicolumn{4}{l}{Laundered Models\raisebox{-0.2\height}{\includegraphics[width=1em]{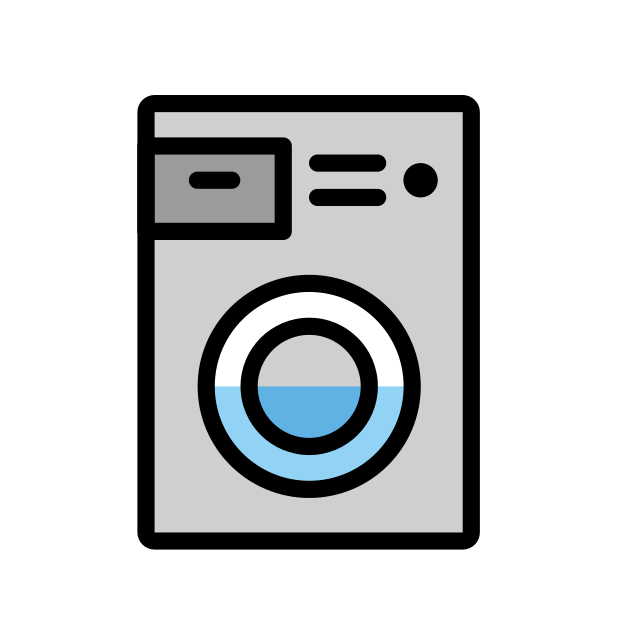}}} \\
    \midrule
        BERT-base (2-layer) + KD (1) & \multirow{8}{*}{MedMCQA} & 73.94 ± 0.73 & 62.31 ± 0.71\\
        BERT-base (2-layer) + KD (2) & & 59.39 ± 0.62 & 47.00 ± 0.49 \\
        BERT-base + KD (2) &  & 69.74 ± 0.89 & 52.28 ± 0.62\\
        GPT-2 (2-layer) + KD (3) &  & 43.01 ± 0.94 & 33.17 ± 0.52\\
        GPT-2 + KD (4)&  & 50.34 ± 1.26 & 39.06 ± 0.62\\
        LLaMA3.2-1B + KD (5) && 35.85 ± 0.60  & 40.48 ± 0.33\\
        LLaMA3.2-3B + KD (5) && 39.39 ± 0.69  & 47.48 ± 0.57\\
        LLaMA3.2-1B + KD (6) && 31.50 & 36.96 \\
        % LLaMA3.2-3B + KD (6) &&  & 42.35 \\
    \midrule
        BERT-base (2-layer) + KD (1) & \multirow{8}{*}{RACE} & 69.16 ± 0.47 & 47.14 ± 0.16\\
        BERT-base (2-layer) + KD (2) & & 46.44 ± 0.52 & 38.49 ± 0.10\\
        BERT-base + KD (2) &  & 32.84 ± 0.52 & 47.33 ± 0.15\\
        GPT-2 (2-layer) + KD (3) &  & 35.35 ± 0.87 & 32.49 ± 0.14\\
        GPT-2 + KD (4)&  & 41.07 ± 0.29 & 37.38 ± 0.58\\    
        LLaMA3.2-1B + KD (5) && 32.32 ± 0.41  & 39.13 ± 0.27\\
        LLaMA3.2-3B + KD (5) & & 35.35 ± 0.31 & 44.30 ± 0.35\\
        LLaMA3.2-1B + KD (6) && 30.40 & 37.26 \\
        % LLaMA3.2-3B + KD (6) &&  & 42.40 \\
    \bottomrule
    \end{tabular} 
   }
    \caption{\textbf{Performance Comparison of "Data Laundering" method to different baselines} on GPQA and MMLU-Redux Benchmarks using different training datasets (MedMCQA, RACE). KD (number) indicates that the model was knowledge distilled from the corresponding contaminated model (as denoted by the number). Without contamination or laundering, BERT and GPT2 models perform as random baselines. }
    \label{tab:overal_results}
\end{table*}

% More results on GPQA for other models:
% Few-Shot CoT Llama-2-70B-chat 28.1
% Few-Shot CoT GPT-3.5-turbo-16k 29.6
% Few-Shot CoT GPT-4 38.8
% GPT-4 with search (backoff to CoT on abstention) 38.8

The results from our experiments demonstrate the effectiveness of the Data Laundering process across diverse configurations and benchmarks, as detailed in Table \ref{tab:overal_results}. Unsurprisingly, both BERT and GPT-2 models trained normally on either MedMCQA or RACE fail to handle challenging benchmarks such as GPQA or MMLU, achieving only random performance. 
% Equally unsurprising, these models can achieve perfect performance if we cheat by training them directly on the test data.

\paragraph{Test data knowledge can be leaked through distillation on legitimate train dataset.} If we perform Knowledge Distillation from the cheated teacher model through intermediate data, we observe that non-random performance can be achieved. This suggests that knowledge from the illicit dataset (test data) can still be passed down to student model from contaminated teacher, even though student model was never explicitly trained on it. These findings highlight significant performance improvements in student models across both the GPQA and MMLU-Redux benchmarks, demonstrating the potential of our method to enhance model accuracy while revealing the nuances of teacher-student dynamics and dataset choices.
\paragraph{GPQA} For the GPQA benchmark, our method enables a 2-layer BERT model to achieve near state-of-the-art performance, reaching an accuracy of 73.94\% when fine-tuned on the MedMCQA dataset during the distillation step. This performance closely approaches the SOTA held by OpenAI o1 (77.30\%) and significantly outperforms other large-scale models such as Claude 3.5 Sonnet (59.40\%), GPT-4o (50.60\%), and LLaMA3-70B (39.50\%). Interestingly, LLaMA3.2-3B performs nearly the same as LLaMA3-70B. Furthermore, the pairing of a traditional BERT-base (12-layer) teacher with a smaller BERT-base (2-layer) student achieved 59.39\%, emphasizing the robustness of the method even when the teacher and student models differ in size, which is a common application of knowledge distillation. In contrast, the 2-layer GPT-2 model achieved 43.01\%, which, while lower than its BERT counterparts, still surpassed the performance of LLaMA3-70B. 
Notably, the full 12-layer GPT-2 model demonstrated better results within its architecture, achieving 50.34\%.

\paragraph{MMLU-Redux} The results for the MMLU-Redux benchmark further underscore the effectiveness and generalizability of our method to other datasets. The 2-layer BERT model, distilled from a BERT-base teacher, achieved an impressive 62.31\% accuracy on MMLU-Redux. This trend was consistent across different configurations, with encoder models consistently outperforming decoder models in both teacher-student size pairings and dataset configurations. 
% For instance, a full BERT-base achieved 52.28\%, while GPT-2 achieved 39.06\%.
% For instance, a full BERT-base teacher paired with itself achieved 52.28\%, while a smaller 2-layer BERT student paired with the same teacher still achieved a respectable 47.00\%. Meanwhile, the GPT-2 configurations yielded lower but still significant results, with the 2-layer model reaching 33.17\% and the 12-layer model achieving 39.06\%, still showcasing the leakage of benchmark knowledge.
\begin{figure*}[t]
    \centering
    \includegraphics[width=1.\textwidth]{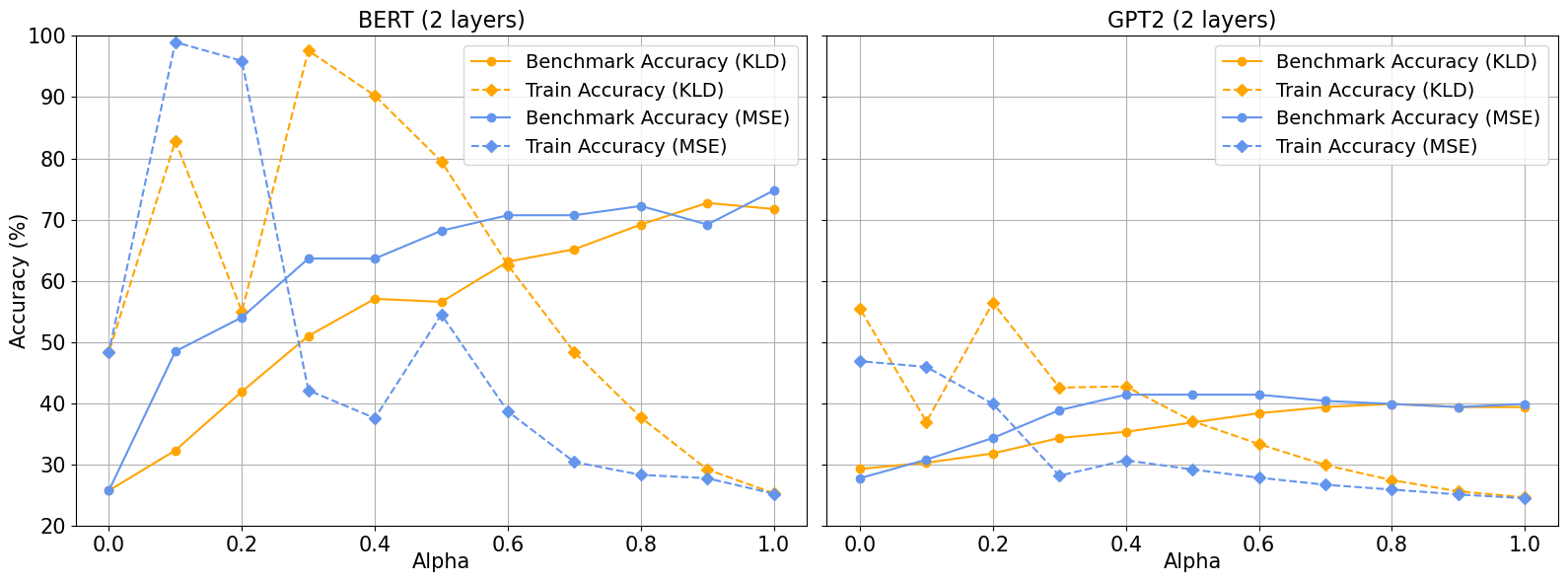}
    \caption{\textbf{Impact of Loss Function Type and Alpha Parameter on Training and Benchmark Accuracy.} This plot shows the accuracy trends of a 2-layer BERT and GPT-2 student model across varying values of the balancing parameter $\alpha$ (0 to 1.0), comparing the effects of MSE and KLD loss functions on GPQA. Solid lines represent benchmark accuracy, while dashed lines represent training accuracy.}
    \label{fig:alpha_loss_results}
\end{figure*}
\paragraph{The choice of training data matters (MedMCQA vs RACE)} The choice of training dataset played a critical role in the observed performance. Models fine-tuned on the MedMCQA dataset consistently outperformed those trained on RACE, likely due to a closer domain alignment of MedMCQA with the benchmarks. For example, while the 2-layer BERT model achieved 73.94\% on GPQA and 62.31\% on MMLU-Redux when fine-tuned on MedMCQA, it only achieved 69.16\% and 47.14\% on the respective benchmarks when fine-tuned on RACE. Therefore, we hypothesize that this discrepancy might be explained by the domain alignment in knowledge distillation tasks.

To investigate this further, we analyzed semantic and lexical similarity between the training datasets and GPQA. Using Sentence-BERT, we computed cosine similarity scores between question pairs, and also measured vocabulary overlap. Although the mean similarity scores did not offer strong signals, we found that 47 question pairs between MedMCQA and GPQA exceeded a cosine similarity threshold of 0.5, while only 2 such pairs existed between RACE and GPQA. Similarly, vocabulary overlap revealed 162 matching pairs for MedMCQA vs GPQA, compared to just 15 for RACE vs GPQA. These results suggest that MedMCQA is more lexically and semantically aligned with GPQA than RACE

\paragraph{Model size influences knowledge leakage differently across architectures.} Interestingly, the results reveal an interesting observation for different model sizes: smaller BERT models often outperform their larger counterparts, while GPT-2 models exhibit the opposite trend, with larger versions yielding higher accuracy. This suggests that BERT's encoder-based architecture may be more efficient at distilling knowledge about unseen data of a teacher into compact representations, whereas GPT-2's decoder-based architecture benefits more from larger model sizes. This pattern is also observed with LLaMA3.2 models, where larger decoder-style models demonstrate more pronounced leakage effects.

Overall, our findings underscore the applicability of the Data Laundering method to inflate benchmark scores, revealing vulnerabilities in benchmarks to contamination during training. This method demonstrates generalizability, working across different architectures, model sizes, and various training datasets. Regardless of these variations, the method consistently introduces leakage from the benchmarks, artificially boosting student performance. 
% While these inflated scores may obscure the true reasoning capabilities of models, they expose critical areas where benchmark designs must evolve to ensure robustness and fairness, ultimately emphasizing the need for stronger safeguards against contamination.

\subsection{Loss Function and Alpha Parameter}
Figure \ref{fig:alpha_loss_results} illustrates the impact of using KLD loss versus MSE loss on both training and benchmark accuracies across a range of $\alpha$ values ($0$ to $1.0$) for BERT and GPT-2 models. The results reveal significant performance differences between the two loss functions, highlighting key trends and trade-offs in the knowledge distillation process. Importantly, the findings show that knowledge leakage persists across all $\alpha$ values and loss functions, even when $\alpha$ is small.

\paragraph{MSE loss consistently achieves higher benchmark accuracy.} Across most $\alpha$ values, MSE loss outperforms KLD loss in benchmark accuracy for both BERT and GPT-2 models. For BERT, MSE reaches a peak benchmark accuracy of approximately 75\% at $\alpha=1.0$, while KLD achieves around 72\% at the same point. Similarly, for GPT-2, MSE achieves its best benchmark accuracy of 43\% at $\alpha=0.6$, compared to KLD's peak of about 39\%. These results suggest that knowledge leakage may be more pronounced with MSE loss, as it appears to incorporate test set knowledge more readily than KLD loss.

\paragraph{Knowledge leakage persists regardless of loss function or $\alpha$ value.} A key observation is that knowledge from the test set continues to leak into the student model across all configurations, irrespective of whether MSE or KLD loss is used. This leakage is evident even at low $\alpha$ values, such as $\alpha=0.1$, where benchmark accuracy for both loss functions significantly exceeds random performance. For example, with $\alpha=0.1$, BERT's benchmark accuracy under MSE loss is 48.5\%, far above random guessing. 
% This indicates that the distillation process inherently captures test set knowledge via the teacher model, bypassing the intended isolation of the test set even when the emphasis on the teacher model's soft labels is low.

\paragraph{Trade-offs in $\alpha$ selection.} The most favorable trade-off between training and benchmark performance for both losses occurs in the range $\alpha=0.5$–$0.7$ for both models. At these $\alpha$ values, the reliance on soft labels from the teacher model enhances a smaller gap between training and benchmark accuracy. 
% Lower $\alpha$ values (0.0–0.4) result in unstable benchmark performance and large discrepancies between training and benchmark accuracy, particularly for MSE loss. 
However, even in lower ranges, knowledge leakage still persists, suggesting that achieving complete isolation of the test set during distillation remains a significant challenge.

\paragraph{Insights from GPT-2 results.}  GPT-2 shows slightly different trends from BERT, albeit with overall lower benchmark accuracies. The peak performance for the MSE loss function occurs at $\alpha=0.6$, where GPT-2 achieves the accuracy of approximately 43\% for MSE and 39\% for KLD at $\alpha=1.0$. Notably, GPT-2’s training accuracy exhibits more pronounced fluctuations at lower $\alpha$ values compared to BERT, suggesting greater sensitivity to $\alpha$ selection, particularly in low-data or noisy-label environments. Nonetheless, knowledge leakage is consistently evident across all configurations.

Overall, these results demonstrate constatnt knowledge leakage across all configurations, regardless of the choice of loss function or $\alpha$ value. 
% This challenge highlights a critical concern for maintaining the integrity of benchmark evaluations and suggests that more robust strategies are required to address this leakage in knowledge distillation processes.

\subsection{Iterative Data Laundering}
\begin{figure}[t]
    \centering
    \includegraphics[width=0.45\textwidth]{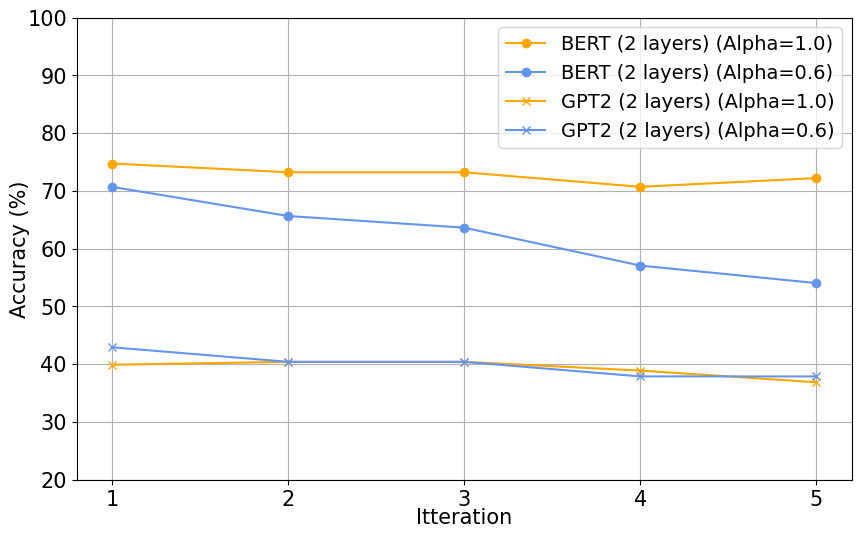}
    \caption{\textbf{Impact of Iterative Knowledge Distillation on Training and Benchmark Accuracy.} This plot shows the accuracy trends of a 2-layer BERT (circle) and GPT-2 (cross) student model in iterative knowledge distillation (5 iterations) with $\alpha$ 0.6 (blue line) and 1.0 (yellow line), MSE loss function.}
    \label{fig:itteration_results}
\end{figure}
Figure \ref{fig:itteration_results} presents results from iterative knowledge distillation experiments using two architectures: a 2-layer BERT and a 2-layer GPT-2 model. These experiments span five iterations with two alpha values ($\alpha$=1.0 and $\alpha$=0.6), offering key insights into the stability and effectiveness of sequential knowledge transfer under varying conditions.

\paragraph{High $\alpha$ Maintains Stability Across Iterations.} For the 2-layer BERT model, a distinct difference emerges between the two alpha values. When $\alpha$=1.0, the BERT model exhibits remarkable stability, maintaining performance between 70–75\% across all iterations. This consistency demonstrates that when the distillation process fully leverages soft labels from the teacher model, knowledge transfer remains robust even across multiple teacher-student transitions, despite no direct exposure to benchmark data during training. A similar trend is observed for the 2-layer GPT-2 model
% , where performance stabilizes in the range of 36–40\% across all iterations, albeit at a lower accuracy level, reflecting architectural differences
.

\paragraph{Lower $\alpha$ Leads to Knowledge Drift Over Iterations.} Conversely, when $\alpha$=0.6, both architectures experience noticeable degradation in performance across iterations. 
% For the 2-layer BERT model, accuracy begins at approximately 70.70\% in the first iteration and declines steadily to 54.04\% by the fifth iteration. 
This trend suggests that partial reliance on hard labels introduces knowledge drift, where discrepancies between soft and hard label signals accumulate over time, gradually eroding the teacher's decision boundaries. Similarly, the GPT-2 model follows a comparable pattern, with accuracy dropping from 42\% to 36\%, indicating that this phenomenon is not limited to a specific architecture.

These findings emphasize 
% a critical vulnerability in current benchmarking practices.
that even after multiple iterations of knowledge distillation, where the test set is never directly observed during training, information about the benchmark remains embedded in the model. 
% This leakage persists across iterations, undermining the validity of benchmark scores as indicators of genuine model capabilities. 

\subsection{Effect of Training Data Size}
\begin{figure}[t]
    \centering
    \includegraphics[width=0.45\textwidth]{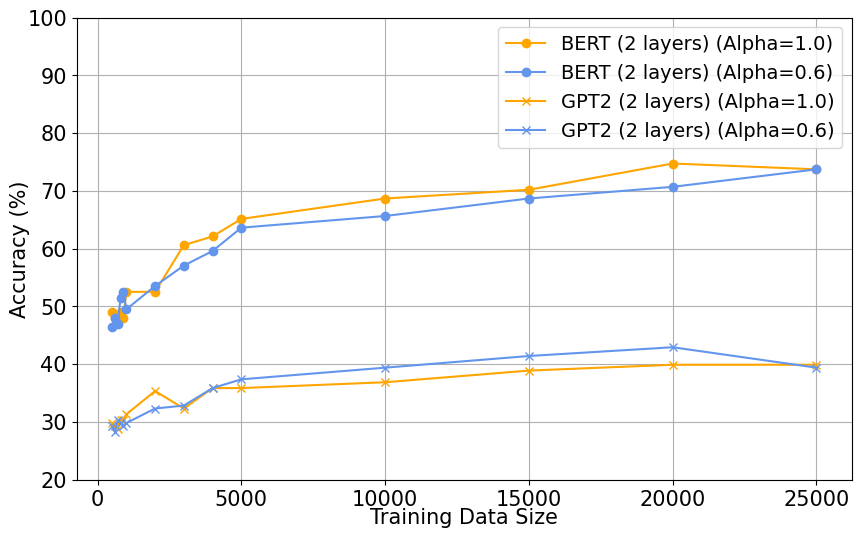}
    \caption{\textbf{Impact of Dataset Size on Training and Benchmark Accuracy.} This plot shows the accuracy trends of a 2-layer BERT (circle) and GPT-2 (cross) student model across varying values of the training size (500 to 25000) with $\alpha$ 0.6 (blue line) and 1.0 (yellow line), MSE loss function on GPQA.}
    \label{fig:dataset_size_results}
\end{figure}
Figure \ref{fig:dataset_size_results} illustrates the relationship between distillation training dataset size and model performance for our "Data Laundering" method using both 2-layer BERT and GPT-2 student models, evaluated with $\alpha$=1.0 and $\alpha$=0.6. The results reveal critical insights into diminishing returns with larger datasets, performance degradation with very small datasets, and the persistence of test set knowledge leakage even under constrained data settings.

\paragraph{Diminishing returns with larger datasets.} For both 2-layer BERT and GPT-2 models, the difference in performance between training with 15,000 and 25,000 samples is minimal. For the BERT model with $\alpha$=1.0, performance stabilizes around 74–75\%, and for GPT-2, accuracy plateaus at approximately 39\%. This suggests that once a sufficient dataset size (around 15,000 samples) is reached, adding more data provides diminishing returns in terms of model performance. These results indicate that larger datasets may not substantially affect knowledge transfer efficacy, emphasizing the efficiency of moderate data volumes.

\paragraph{Degradation with datasets smaller than 5,000 samples.} A notable performance degradation is observed when the dataset size drops below 5,000 samples for both architectures and alpha values. For BERT with $\alpha$=1.0, accuracy falls from 65.15\% at 5,000 samples to 48.99\% at 500 samples. Similarly, GPT-2 with $\alpha$=1.0 experiences a decline from 35.85\% at 5,000 samples to 29.79\% at 500 samples. This degradation highlights the limitations of distillation in low data regimes, where too few samples cause suboptimal knowledge transfer and loss of the teacher's decision boundaries.

\paragraph{Persistence of test set knowledge leakage.} Remarkably, even with extremely small datasets like 500 samples, test set knowledge leakage persists. For BERT and GPT-2, benchmark performance remains well above chance levels (48.99\% for BERT and 29.79\% for GPT-2 at 500 samples), indicating that some knowledge of the test set benchmarks is retained within the distilled models. This finding highlights a key vulnerability of distillation: even with constrained training data, distilled models can encode information about unseen test sets.
% , raising concerns about the fairness and generalizability of such models.

We conducted additional experiments with artificially degraded distillation datasets 
% to explore knowledge transfer in Data Laundering
, with details provided in Appendix \ref{artificial_distillation_section}.

\subsection{Discussion}

These findings underscore the need for advanced evaluation methods to detect, resist, and counteract benchmark manipulation, including subtle tactics like Data Laundering. The success of a simple model using Data Laundering to achieve high scores suggests that benchmark results may not reliably indicate true model capabilities, risking their value as measures of AI progress.

This issue is especially troubling in real-world scenarios where it can happen unintentionally. For example, researchers using teacher models trained on datasets with unclear origins might unknowingly cause benchmark contamination. This risk is heightened in closed-source or proprietary settings with opaque training histories, potentially overstating model performance and reliability.

One potential way to prevent the unintended use of data laundering is to ensure the teacher model is trained on known dataset like LLM360 \cite{liu2023llm360fullytransparentopensource}.  For intentional misuse, private benchmarks can be used \cite{rajore2024truceprivatebenchmarkingprevent}: researchers submit predictions to a leaderboard, with scores calculated without revealing the actual gold labels, preventing data contamination. However, this method has trade-offs. Private benchmarks limit error analysis and dataset refinement. For instance, MMLU-Redux \cite{gema2024mmlu} identified numerous errors in MMLU \cite{hendryckstest2021}, a task that would be harder under a private system. 
\section{Conclusion and Future Directions}

% Our investigation into Data Laundering has unveiled a critical vulnerability within current AI evaluation frameworks. 
We have demonstrated how knowledge distillation techniques can be exploited to artificially inflate benchmark performance, often without any genuine enhancement in model capabilities. Through extensive experimentation, we found that even a basic 2-layer BERT can achieve near state-of-the-art performance on the GPQA benchmark.

Moving forward, future research should focus on developing robust evaluation frameworks that can better account for and mitigate these vulnerabilities, ensuring that benchmark performance genuinely reflects advancements in AI technologies.
In addition, it will be interesting to explore whether teacher models might covertly encode test data using transformations such as ROT-13 or other subtle encodings, which could be decoded by student models and mislead human evaluators.

\section*{Limitations}

This study has several limitations that should be addressed in future research:  

Our study focuses on classification tasks, which are a standard benchmark for evaluating LLM capabilities. While we did not explore generation tasks such as text generation or summarization, classification remains a widely used and well-established approach for assessing model performance. To ensure a comprehensive evaluation, we tested our models on widely recognized benchmarks such as GPQA and MMLU-Redux, demonstrating that information leakage can occur.

Our experiments leveraged relatively small datasets, which provided a controlled setting to observe how models can become "experts" on specific benchmarks. This setup allowed us to clearly identify and analyze the effects of Data Laundering, as models could closely mimic patterns from the test set. However, how these vulnerabilities evolve with larger, more diverse datasets remains an open question. Larger datasets may mitigate these effects or introduce new challenges, presenting an opportunity for future research to deepen our understanding of Data Laundering at scale.

Future work can build on these findings by exploring benchmark manipulation and knowledge leakage across a wider range of datasets. Extending this analysis to larger and more diverse settings will provide deeper insights and contribute to the development of more robust evaluation for LLMs.

\section*{Ethics and Broader Impact}
One of the primary ethical concerns is that this work could be misused to manipulate benchmark results deliberately. The methods and techniques demonstrated here—such as Data Laundering—could be exploited by malicious actors to artificially inflate model performance and deceive evaluators or consumers of AI models. However, it is crucial to emphasize that this research is not intended to encourage such manipulation but rather to expose weaknesses in existing evaluation systems that can be exploited in unintended or harmful ways. Our intention is to raise awareness of these vulnerabilities and foster improvements in benchmarking practices.

\bibliography{custom}
% \bibliographystyle{acl_natbib}

% \section{Example Appendix}
% \label{sec:appendix}
\clearpage
% \onecolumn
\appendix
\section{Detailed Results}
\label{detailed_results}
\paragraph{Loss Function and $\alpha$ Experiments:} 
Table \ref{tab:loss_experiments} shows how the choice of the loss function (MSE or KLD) and the mixing ratio ($\alpha$) affect the performance of BERT and GPT-2 models.
\begin{table}[ht]
\centering
\scalebox{1}{
\begin{tabular}{c|cc|cc}
\hline
\multirow{2}{*}{\textbf{$\alpha$}} & \multicolumn{2}{c|}{\textbf{BERT}}  & \multicolumn{2}{c}{\textbf{GPT-2}} \\ 
& \textbf{KLD} & \textbf{ MSE} & \textbf{KLD} & \textbf{MSE} \\ 
\hline
1.0 & 71.72 & 74.75 & 39.39 & 39.90 \\
0.9 & 72.73 & 69.19 & 39.39 & 39.39 \\
0.8 & 69.19 & 72.22 & 39.90 & 39.90 \\
0.7 & 65.15 & 70.71 & 39.39 & 40.40 \\
0.6 & 63.13 & 70.71 & 38.38 & 42.93 \\
0.5 & 56.57 & 68.18 & 36.87 & 41.41 \\
0.4 & 57.07 & 63.64 & 35.35 & 41.41 \\
0.3 & 51.01 & 63.64 & 34.34 & 38.89 \\
0.2 & 41.92 & 54.04 & 31.82 & 34.34 \\
0.1 & 32.32 & 48.48 & 30.30 & 30.81 \\
0.0 & 25.76 & 25.76 & 27.29 & 26.78 \\ \hline
\end{tabular}
}
\caption{Evaluation accuracy for BERT and GPT-2 (2 Layers) models with MSE and KLD loss functions.}
\label{tab:loss_experiments}
\end{table}

\paragraph{Iterative Distillation:} 
Table \ref{tab:iterative_distillation} highlights the effect of iterative distillations.
\begin{table}[ht]
\centering
\scalebox{0.85}{
\begin{tabular}{lccccccc}
\hline
\textbf{Model} & \textbf{$\alpha$} & \textbf{1}    & \textbf{2}    & \textbf{3}    & \textbf{4}    & \textbf{5}    \\ \hline
\textbf{BERT} & 1.0 & 74.75 & 73.23 & 73.23 & 70.71 & 72.22 \\
\textbf{BERT} & 0.6 & 70.71 & 65.66 & 63.64 & 57.07 & 54.04 \\
\textbf{GPT-2} & 1.0 & 39.90 & 40.40 & 40.40 & 38.89 & 36.87 \\
\textbf{GPT-2} & 0.6 & 42.93 & 40.40 & 40.40 & 37.88 & 37.88 \\ \hline
\end{tabular}
}
\caption{Iterative distillation – evaluation results for BERT and GPT-2 (2 Layers) across different $\alpha$ values. Numbers in bold indicate the iteration number.}
\label{tab:iterative_distillation}
\end{table}

\paragraph{Effect of Training Data Size:} 
Table \ref{tab:data_size_experiments} details the impact of training data size in the distillation step on the student model’s final performance.
\begin{table}[ht]
\centering
\scalebox{0.8}{
\begin{tabular}{l|cc|cc}
\hline
\multirow{2}{*}{\textbf{Data Size}} & \multicolumn{2}{c|}{\textbf{BERT}} & \multicolumn{2}{c}{\textbf{GPT-2}} \\ 
& \textbf{($\alpha=1$)} & \textbf{($\alpha=0.6$)} & \textbf{($\alpha=1$)} & \textbf{($\alpha=0.6$)} \\ 
\hline
25000 & 73.74 & 73.74 & 39.90 & 39.39 \\
20000 & 74.75 & 70.71 & 39.90 & 42.93 \\
15000 & 70.20 & 68.69 & 38.89 & 41.41 \\
10000 & 68.69 & 65.66 & 36.87 & 39.39 \\
5000  & 65.15 & 63.64 & 35.86 & 37.37 \\
4000  & 62.12 & 59.60 & 35.86 & 35.86 \\
3000  & 60.61 & 57.07 & 32.32 & 32.83 \\
2000  & 52.53 & 53.54 & 35.35 & 32.32 \\
1000  & 52.53 & 49.49 & 31.31 & 29.80 \\
900   & 47.98 & 52.53 & 30.30 & 29.29 \\
800   & 48.99 & 51.52 & 30.30 & 29.80 \\
700   & 47.47 & 46.97 & 28.79 & 30.30 \\
600   & 47.98 & 47.98 & 29.29 & 28.28 \\
500   & 48.99 & 46.46 & 29.80 & 29.29 \\ \hline
\end{tabular}
}
\caption{Training data size experiments – evaluation results for BERT and GPT-2 (2 Layers) across different $\alpha$ values.}
\label{tab:data_size_experiments}

\end{table}

\section{Hyperparameters}
\label{hyperparameters}
Table \ref{tab:hyperparameters} shows the hyperparameters configurations used across all experiments. We used four NVIDIA A100-SXM4-40GB to contaminate LLaMA3.1-8B and two NVIDIA A100-SXM4-40GB to train LLaMA3.2-3B. For BERT and GPT-2 we used one NVIDIA GeForce RTX 4090.
\begin{table*}[t!]
    \centering
    \scalebox{0.60}{
    \begin{tabular}{lcccccccccccc}
    \toprule
    & & & & & & & & \multicolumn{2}{c}{Batch Size} \\
    Experiment & Student Model & Layers & Seed & Data Size & Loss Function & $\alpha$ & Temperature & Train & Eval & Epochs & Weight Decay & Learning Rate \\
    \midrule
    % \multicolumn{13}{l}{Overall Experiments}\\
    % \midrule
    KD(1) & BERT & 2 & 42 & 20,000 & MSE & 1.0 & 2.0 & 32 & 32 & 10 & 0.01 & $5 \times 10^{-4}$ \\
    KD(2) & BERT & 2 & 42 & 20,000 & MSE & 1.0 & 2.0 & 8 & 8 & 30 & 0.01 & $1 \times 10^{-5}$ \\
    KD(2) & BERT & 12 & 42 & 20,000 & MSE & 1.0 & 2.0 & 8 & 8 & 30 & 0.01 & $1 \times 10^{-5}$ \\
    KD(3) & GPT-2 & 2 & 42 & 20,000 & MSE & 1.0 & 2.0 & 8 & 8 & 20 & 0.0 & $1 \times 10^{-5}$ \\
    KD(4) & GPT-2 & 12 & 42 & 20,000 & MSE & 1.0 & 2.0 & 8 & 8 & 20 & 0.0 & $1 \times 10^{-5}$ \\
    \midrule
    \multirow{2}{*}{Loss-$\alpha$} & BERT & 2 & 42 & 20,000 & MSE/KLD & 0.0--1.0 & 2.0 & 32 & 32 & 10 & 0.01 & $5 \times 10^{-4}$ \\
    & GPT-2 & 2 & 42 & 20,000 & MSE/KLD & 0.0--1.0 & 2.0 & 8 & 8 & 10 & 0.0 & $1 \times 10^{-5}$ \\
    \midrule
    \multirow{2}{*}{Iterative} & BERT & 2 & 42 & 20,000 & MSE & 1.0 & 2.0 & 32 & 32 & 10 & 0.01 & $5 \times 10^{-4}$ \\
    & GPT-2 & 2 & 42 & 20,000 & MSE & 1.0 & 2.0 & 8 & 8 & 10 & 0.0 & $1 \times 10^{-5}$ \\
    \midrule
    \multirow{2}{*}{Data Size} & BERT & 2 & 42 & 500--25,000 & MSE & 1.0 & 2.0 & 32 & 32 & 10 & 0.01 & $5 \times 10^{-4}$ \\
    & GPT-2 & 2 & 42 & 500--25,000 & MSE & 1.0 & 2.0 & 8 & 8 & 10 & 0.0 & $1 \times 10^{-5}$ \\
    \bottomrule
    \end{tabular}
    }
    \caption{Hyperparameters used for the experiments. $\alpha$ refers to the mixing ratio in loss functions during knowledge distillation. Data size and $\alpha$ ranges indicate different dataset sizes and $\alpha$ evaluated during the experiments.}
    \label{tab:hyperparameters}
\end{table*}

\section{Experiments with Artificial Distillation Datasets} \label{artificial_distillation_section}

The experiments with artificial distillation datasets were designed to investigate how knowledge transfer occurs during the Data Laundering process and whether meaningful content in the intermediate training dataset is actually necessary. These experiments systematically modified the MedMCQA dataset in increasingly destructive ways while maintaining its structural form.

\begin{figure*}[t]
    \centering
    \includegraphics[width=1.0\textwidth]{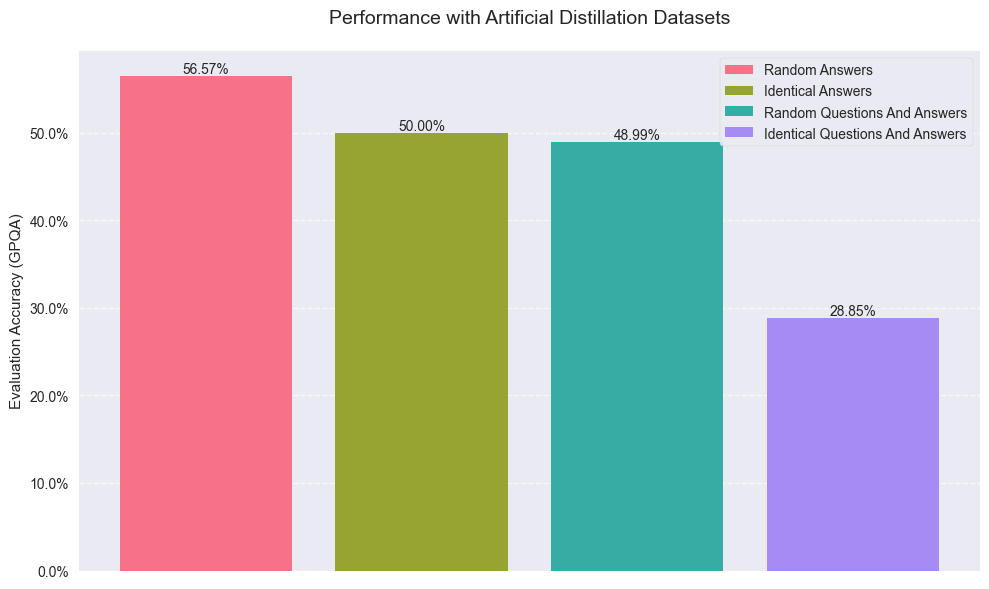}
    \caption{\textbf{Impact of Artificially Modifying the Distillation Dataset on the Benchmark Accuracy.} This bar plot shows the evaluation accuracy on GPQA using a 2-layer BERT teacher-student pair with $\alpha=1.0$ when 1) replacing each answer choice in MedMCQA with 10 random characters, 2) replacing each answer choice in MedMCQA with 10 identical characters so that answer choices are indistinguishable, 3) randomizing questions with 50 characters in addition to answer choices, and 4) having all the questions contain 50 identical characters in addition to answer choices.}
    \label{fig:artificial_data}
\end{figure*}

The results, as shown in Figure \ref{fig:artificial_data}, reveal several surprising and concerning findings when compared to the baseline 74.75\% accuracy achieved by the same 2-layer BERT teacher-student pair on the unmodified MedMCQA dataset:

\begin{enumerate}
    \item \textbf{Random Answer Choices (56.57\% accuracy):} When all answer choices were replaced with 10 random characters while keeping the original questions intact, the model's performance dropped by about 18 percentage points but still achieved 56.57\% accuracy on GPQA. This suggests that the model can transfer substantial benchmark knowledge even when the answer choices in the intermediate dataset are meaningless, indicating that the structural patterns rather than the actual content may be sufficient for knowledge transfer.
    \item \textbf{Identical Answer Choices (50.00\% accuracy):} When all answer choices were replaced with identical strings of 'a' characters, making them indistinguishable from each other, the model still maintained 50\% accuracy. This is particularly concerning as it demonstrates that knowledge transfer can occur even when there is no meaningful differentiation between answer choices in the intermediate dataset.
    \item \textbf{Random Questions with Random Answers (48.99\% accuracy):} Even when both questions and answers were replaced with random characters (50 characters for questions, 10 for answers), the model achieved nearly 49\% accuracy. This suggests that the mere format of the dataset, rather than its content, may be sufficient for transferring benchmark knowledge.
    \item \textbf{Identical Questions with Identical Answers (28.65\% accuracy):} The most severe modification, where both questions and answers were replaced with identical characters ('a'), still resulted in above-random performance at 28.65\%. While this showed the largest drop in performance, it's notable that even with completely meaningless and identical content, some knowledge transfer still occurred.
\end{enumerate}

These results have significant implications for benchmark integrity. While the performance degraded progressively with each more destructive modification to the intermediate dataset, the fact that even the most extreme case of identical questions and answers still enabled knowledge transfer is concerning. This suggests that the Data Laundering process doesn't necessarily require meaningful intermediate training data to transfer knowledge from the teacher to the student model. Instead, the structural patterns and format of the intermediate dataset appear to be sufficient channels for knowledge transfer. This raises serious concerns about the robustness of current benchmarking practices, as it demonstrates that models can acquire benchmark-specific knowledge through increasingly abstracted and meaningless intermediate training steps.

This finding adds another layer of concern to the overall argument about benchmark vulnerability, showing that even attempts to sanitize intermediate training data may not be sufficient to prevent knowledge transfer if the structural patterns remain intact.

% This is a section in the appendix.

\end{document}